\documentclass[lettersize,journal]{IEEEtran}
\usepackage{amsmath,amsfonts}
\usepackage{algorithmic}
\usepackage{algorithm}
\usepackage{array}
\usepackage[caption=false,font=normalsize,labelfont=sf,textfont=sf]{subfig}
\usepackage{textcomp}
\usepackage{stfloats}
\usepackage{url}
\usepackage{verbatim}
\usepackage{graphicx}
\usepackage{cite}
\usepackage{booktabs}

\usepackage{tikz-cd}

\usepackage{amsmath,amssymb,amsthm,mathrsfs,amsfonts,dsfont}

\theoremstyle{theorem} 
\theoremstyle{definition} 
\theoremstyle{definition} 
\theoremstyle{remark} 
\theoremstyle{remark} 
\theoremstyle{theorem} 
\begin{document}

\title{Self-degraded contrastive domain adaptation for industrial fault diagnosis with bi-imbalanced data}

\author{Gecheng Chen, Zeyu Yang, Chengwen Luo, Jianqiang Li
}

\markboth{IEEE Transactions on neural networks and learning systems}%
{Shell \MakeLowercase{\textit{et al.}}: A Sample Article Using IEEEtran.cls for IEEE Journals}


\maketitle

\begin{abstract}
Modern industrial fault diagnosis tasks often face the combined challenge of distribution discrepancy and bi-imbalance. Existing domain adaptation approaches pay little attention to the prevailing bi-imbalance, leading to poor domain adaptation performance or even negative transfer. In this work, we propose a self-degraded contrastive domain adaptation (Sd-CDA) diagnosis framework to handle the domain discrepancy under the bi-imbalanced data. It first pre-trains the feature extractor via imbalance-aware contrastive learning based on model pruning to learn the feature representation efficiently in a self-supervised manner. Then it forces the samples away from the domain boundary based on supervised contrastive domain adversarial learning (SupCon-DA) and ensures the features generated by the feature extractor are discriminative enough. Furthermore, we propose the pruned contrastive domain adversarial learning (PSupCon-DA) to pay automatically re-weighted attention to the minorities to enhance the performance towards bi-imbalanced data. We show the superiority of the proposed method via two experiments.
\end{abstract}


\section{Introduction}
As modern industries have grown in complexity and scale, there has been a significant increase in attention to precise and efficient data-driven fault diagnosis methods to guarantee reliable and stable operations. These methods are designed to identify various fault conditions in processes or machines using extensive data collected from sensors. Recently, deep learning (DL) techniques, such as fully connected neural networks (FNN), and convolutional neural networks (CNN), have been widely applied to data-driven diagnosis and achieved distinguished performance due to their feature extraction ability and end-to-end structures \cite{yang2022paradigm,wu2018deep,song2018fault}. 

However, traditional DL methods only work well under a strong assumption, that is, the training data (denoted as source domain) and the testing data (denoted as target domain) should have the same distribution, which can hardly hold in real-world industrial applications. For example, in an industrial production process, the training data may be gathered under specific working conditions but the condition of interest can be another one with a different working load. As a result, traditional DL-based fault diagnosis methods may fail because these two conditions have different data distributions. 

\begin{figure*}[ht]
    \centering
    \includegraphics[width=0.9\textwidth]{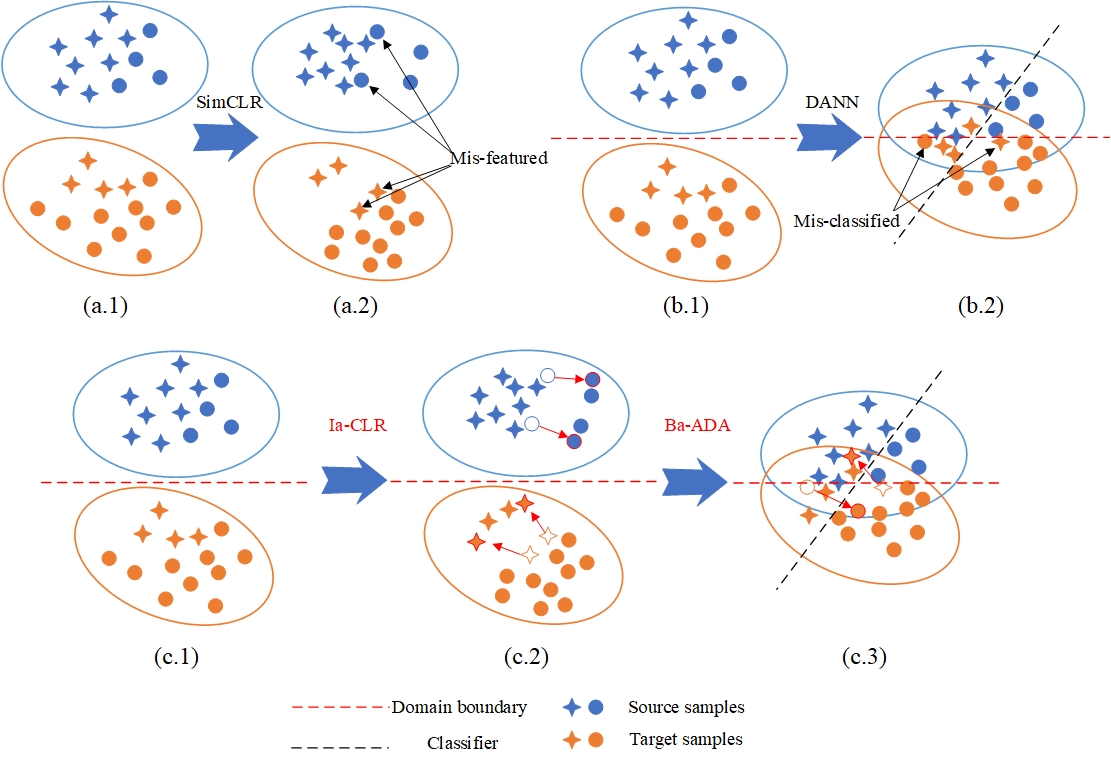}
    \caption{The intuition behind the proposed method. (a.1)-(a.2) illustrates the process of SimCLR, (b.1)-(b.2) shows the process of DANN and (c.1)-(c.3) introduces what the proposed Sd-CDA does.}
    \label{fig:intuition}
\end{figure*}

On the basis of domain discrepancy, however, the industrial data also exhibits heavy imbalance because the processes mainly run in normal conditions. In the scenario of domain discrepancy, the impact of imbalance is two-fold: \textit{intra-domain} imbalance and \textit{inter-domain} imbalance \cite{ding2023deep}. We use (a.1) in Figure \ref{fig:intuition} to illustrate these two imbalances. In (a.1), blue and yellow samples represent samples from two domains, separately, whereas circles and stars represent two different classes. Intra-domain imbalance means that the amount of the normal class (majority) can be much larger than the fault classes (minorities) in both domains.
For example, the number of blue stars is much larger than circles. Inter-domain imbalance indicates that the probabilities of different faults of two domains can be highly different, leading to different imbalance ratios of these two domains. In (a.1) we can see that the circles are the minority class in the blue domain but they are the majority class in the yellow domain. 
We call the data with two kinds of imbalance \textit{bi-imbalanced} data. 

In order to solve the domain discrepancy, the unsupervised domain adaptation (UDA) has achieved great progress \cite{li2020systematic,chen2023transfer}. It leverages the knowledge of the label-rich source domain to solve a similar task on a different but related target domain with no label. Lots of metric-based UDA approaches have been proposed to minimize the discrepancy measured by some explicit function to align these two domains, e.g., maximum mean discrepancy (MMD) \cite{gretton2012kernel} or correlation alignment \cite{sun2016deep}. 
However, the pre-selected metric function may not always be applicable, impeding the applications of metric discrepancy-based DA approaches. Recently, domain adversarial neural networks (DANN) \cite{ganin2016domain} and related techniques have received increasing attention because they measure the discrepancy between two domains implicitly by a domain discriminator. In general, domain adversarial learning-based approaches train a feature extractor and a domain discriminator adversarially to extract domain invariant features from the two domains and construct an efficient classifier for the target domain. However, a model well-trained based on source labels may overfit the source domain and cannot generalize well to the target domain \cite{chai2020fine}. Even worse, the domain adversarial learning-based methods are not guaranteed to generate discriminative features for all samples. The main reason is that the feature extractor can simply generate features along the domain boundary because the goal of the feature extractor is to fool the domain discriminator during training \cite{saito2018maximum}. When the imbalance exists, the minorities can be hardly corrected once they are along the domain boundary due to the coverage of the majority.

Recently, several works have explored self-supervised techniques to improve the representation learning of DANN-related work and increase the diagnosis accuracy in a pre-train manner. Contrastive learning,  originally proposed to extract representation features from images or videos, learns effective representations in the latent space in an unsupervised manner by enforcing the similarity of positive pairs and enlarging the distance of negative pairs \cite{noroozi2016unsupervised,bachman2019learning}. As one of the most popular contrastive learning methods, simple contrastive representation learning (SimCLR) \cite{chen2020simple} has been introduced to domain adaptation methods for industrial fault diagnosis \cite{lu2022domain,azuma2023adversarial}. However, this contrastive learning-based approach is highly vulnerable to imbalanced data and may make some minority samples mis-featured. The reason is that the contrastive learning of minorities can be covered by the majorities under imbalanced settings because the instance-rich class has much more separable information \cite{jiang2021self}.


In \cite{hooker2019compressed}, the authors prove that DNNs tend to memorize the majorities but forget the minorities or difficult samples after pruning or compression. In other words, for a minority sample or a sample along the decision boundary, the difference between extracted features given by a DNN and its pruned version should be larger compared with a normal sample. 
Inspired by this, we propose a novel domain adaptation framework called Self-degraded Contrastive Domain adaptation (Sd-CDA) for the aforementioned two challenges, which consists of two parts: imbalance-aware contrastive learning for representations (Ia-CLR) and boundary-aware adversarial domain adaptation (Ba-ADA). 
First, to relieve the impact of bi-imbalance on contrastive learning, we propose the imbalance-aware contrastive learning for representations (Ia-CLR) as the pre-train procedure before domain adaptation. Since the original SimCLR learns the feature representations via input augmentation, we can consider it as contrastive learning at a data level. In this work, the proposed Ia-CLR facilitates SimCLR with a model-level view by minimizing the difference between the feature extractor and its pruned counterpart which we can consider as a perturbed version of the original model. Given that the minorities cause more significant differences than the majority, the Ia-CLR will attach higher weights automatically and implicitly to the minorities during the self-supervised training. 
Second, in order to solve the bi-imbalance challenge during the adversarial training and ensure the extraction of discriminative features, the proposed Ba-ADA first introduces the idea of supervised contrastive learning to the discriminator of DANN to make the extracted features of two domains more compact. By doing this, it forces the feature extractor to generate discriminative features away from the domain boundary. What's more, Ba-ADA proposes a pruned supervised contrastive learning to help detect the minorities of the target domain along the domain boundary by pruning the discriminator. In summary, the main innovations of this work are as follows:
\begin{itemize}
    \item Propose an imbalance-aware contrastive pre-train approach by pruning the feature extractor to enhance the representation learning of the bi-imbalance data on the basis of traditional constrastive learning.
    \item Introduce the idea of supervised contrastive learning to DANN to force the feature extractor to generate discriminative features.
    \item Construct a pruned supervised contrastive learning strategy for the domain discriminator to pay more attention to the minorities under bi-imbalance during the adversarial training.
\end{itemize}

The remainder of this paper is organized as follows. We first review several related works in Section \ref{related}. Then we introduce the proposed Sd-CDA framework in detail in Section \ref{method}. In Section \ref{experiments} we demonstrate the superiority of the proposed method via two cases. Conclusion remarks and discussion are given in Section \ref{conclusion}. 



\section{Related work}\label{related}
\subsection{Unsupervised domain adaptation}
Current unsupervised domain adaptation approaches for industrial fault diagnosis can be mainly divided into two categories: metric-based methods and adversarial learning-based methods. 
Metric-based domain adaptation methods mainly measure the discrepancy between the source and the target domain based on several explicit metric functions, e.g., MMD, and correlation alignment (CORAL). Lu \cite{lu2016deep} proposes a domain adaptation fault diagnosis method to reduce the MMD between the source and the target domains. Jiao \cite{jiao2020residual} introduces the joint MMD to learn transferable features and formulates a residual domain adaptation framework. Li \cite{li2021central} proposes a distribution alignment method to reduce the difference between the two domains based on the central moment metric. However, the pre-selected metric function may not always be applicable and may result in significant computational costs for the neural networks.

Adversarial learning-based methods primarily achieve knowledge transfer by implicitly measuring the difference between different domains and guiding the model to learn domain invariant features based on the idea of Generative Adversarial Networks (GANs). Ganin proposes the Domain Adversarial Neural Networks (DANN) framework, introducing adversarial learning into domain adaptation \cite{ganin2016domain}. The main structure of DANN includes three parts: a feature generator $G$, a domain discriminator $D$ and a label classifier $C$. During the training, $G$ and $C$ are trained jointly to minimize the cross-entropy loss on the source samples with labels. Additionally, $G$ and $D$ are trained adversarially. $D$ tries to discriminate correctly whether a feature generated by $G$ is from source or target sample while $G$ tries to fool $D$. By this adversarial training manner, DANN decreases the discrepancy of the two domains and extracts the domain invariant features for downstream tasks. Chen applies DANN to the task of diagnosing rotating machinery faults \cite{chen2020domain}. Wang uses Wasserstein GAN to make the adversarial learning process smoother [30]. Chai extends the original DANN into a fine-grained version \cite{chai2020fine}. Huang builds a domain adaptation network based on conditional adversarial learning \cite{huang2021multisource}. However, the generated features are not guaranteed to be discriminative enough even if the discriminator is confused. To solve this problem, several recent works rely on pseudo-labels to ensure discriminative features. Lu assigns generated features pseudo-labels based on clustering and corrects the predicted class labels based on the pseudo-labels \cite{lu2022domain}. Jiao formulates a bi-classifier structure to generate high-confidence pseudo-labels for the target samples \cite{jiao2022self}. However, since the pseudo-labels are generated by the feature generator or label classifier given by previous steps, they are not reliable enough and may cause the training to diverge. Compared with the existing approaches, the proposed Sd-CDA forces the feature extractor to generate discriminative features without any pseudo-labels.


\subsection{Imbalanced domain adaptation}
Current imbalanced domain adaptation methods mainly consider the intra-domain imbalance. Based on the idea of sample generation, Zareapoor trains an adversarial generative network model to generate minority class data and diagnostic decisions simultaneously \cite{zareapoor2021oversampling}. However, since the information of minorities is very limited, the generated data may not match the real fault distribution or even cause overfitting. Based on the idea of re-weighting, Xiao combines TrAdaBoost to dynamically weight each sample point according to the classification error of each source domain sample to address the with-domain imbalance \cite{xiao2019transfer}. Based on the principle of feature transfer, Kuang re-weights the loss of each class based on the imbalance ratio when aligning the distributions of different domains \cite{kuang2021class}. Liu explored the sample weights based on meta-learning \cite{liu2022cross}. However, as the target domain distribution is unknown, the resampling methods derived from the source domain imbalance are often not suitable for the target domain. In order to solve the inter-domain imbalance, Wu \cite{wu2022imbalanced} and Ding \cite{ding2023deep} align the features of every class separately via local MMD to avoid the coverage of the majority. However, they still rely on the unreliable pseudo-labels of the target samples to train the model. Compared with these methods, the proposed model assigns weights to the minorities automatically based on whether the model learns them well without any specific functions or pseudo-labels. By doing this, the resulting weights are more reliable and convenient to achieve.

\subsection{Contrastive learning}
Deep learning relies heavily on substantial amounts of labeled training data. However, data annotation can be costly, leading to significant interest in self-supervised learning. Contrastive learning, an emerging self-supervised technique, shows great promise in learning effective representations without the need for label information \cite{noroozi2016unsupervised,bachman2019learning}. It learns effective visual representations in the latent space in an unsupervised manner by enforcing the similarity of positive pairs and enlarging the distance of negative pairs to learn the visual representations.  Simple contrastive representation learning (SimCLR), one of the most prevalent contrastive learning methods, has attracted a lot of attention \cite{chen2020simple}. Assume an unlabeled dataset $X$ whose data size is $N$, SimCLR first augments it via two different augmentations and results in $2N$ data points. For each augmented input, only the other version from the same input is set to be the positive pair, and all the remaining $2(N-1)$ samples are regarded as the negative pairs. The ultimate goal of SimCLR is to minimize the Normalized Temperature-Scaled Cross-Entropy loss (NT-Xent) which is defined as:
\begin{equation}\label{ntloss}
    \mathcal{L}_g=\frac{1}{N}\sum_{i=1}^N-\log \frac{d(v_i,v_i^+)}{d(v_i,v_i^+) + \sum_{v_i^-\in V^-}d(v_i,v_i^-)},  
\end{equation}
where $v_i,v_i^+$ represents the projections of positive pairs, $v_i,v_i^-$ represents projections of negative pairs and $V^-$ means the set of projections of negative pairs. $d(\cdot,\cdot)=\exp (sim(\cdot,\cdot)/\tau)$ calculates the similarity between two projections, $sim(\cdot,\cdot)$ represents a kernel function and $\tau$ is the temperature parameter. On the basis of pure unsupervised contrastive learning, Khosla proposes supervised contrastive learning (SupCon) by considering all samples belonging to the same categories as positive against the negatives from the remainder categories \cite{khosla2020supervised}. SupCon makes the samples of different categories more compact by taking label information into account.

Recently, SimCLR has been introduced to domain adaptation approaches as a pre-train step to avoid the overfitting of DANN towards the source domain. Lu takes different transformed methods as pretext to pre-train the DANN model via contrastive learning \cite{lu2022domain}. Azuma combines SimCLR with DANN and proposes the contrastive domain adaptation framework \cite{azuma2023adversarial}.  However, contrastive learning of minorities can be covered by the majorities under imbalanced settings because the instance-rich class has much more separable information \cite{jiang2021self}. Compared with the existing contrastive learning-based pre-train approaches, the proposed method utilizes the difference between the original model and its pruned counterpart to re-weight the samples implicitly without label information.

\section{Methodologies}\label{method}

\subsection{Motivations and intuition}
We use the SimCLR and DANN as examples to illustrate the impact of bi-imbalance on contrastive learning-based pre-train methods and domain adversarial learning in detail and introduce the main intuition of this work in Figure \ref{fig:intuition}. 

The original SimCLR approach learns effective representations by maximizing the agreement between different versions of augmented unlabeled input. Such pre-train procedure helps the model learn effective feature extraction and achieve fast and stable downstream convergence as shown in (a.1) - (a.2). However this contrastive learning-based approach is highly vulnerable to imbalanced data and may make some minority samples mis-featured as shown in (a.2), and thus influence the downstream training. The existing approaches for the imbalance problem rely on the label information to re-weight samples or generate minorities. However, due to the inter-domain imbalance, the imbalance of the target domain is unknown and cannot be estimated by the imbalance ratio of the source domain. Based on this, a big challenge of implementing effective contrastive learning for industrial fault diagnosis is how to solve the possible imbalance in the data to avoid mis-featured samples without label information of the target domain. Given that the pruned version of a neural network tends to forget the minorities \cite{hooker2019compressed}, it is a natural intuition to use a pruned model to detect the minorities of both domains without the label information as shown in the step (c.1) - (c.2).

In (b.1) - (b.2), we show the general steps of DANN: it decreases the discrepancy between two domains to extract domain invariant features, and then uses the classifier trained on the labeled source domain to classify the target samples. However, the feature extractor can simply generate features along the domain boundary because the goal of the feature extractor is to fool the domain discriminator during training. What's worse, the minorities can be hardly corrected once they are along the domain boundary due to the coverage of the majority when the imbalance exists. As for the misclassified source samples, we can tune the sample weights of minorities to pay more attention to the classification-related features of the minorities. However, how to force the minorities of the target domain away from the domain boundary under bi-imbalanced scenarios remains a big challenge. Since supervised contrastive learning has the ability to make the samples of different categories more compact, it holds the potential to help the domain discriminator drag the samples along the domain boundary back. In order to handle the bi-imbalance and enhance the detection of minorities of the target domain, we can also implement the pruned version of the domain discriminator to help the training as shown in (c.2) - (c.3).

\subsection{Problem formulation and overall structure}
\begin{figure}[ht]
    \centering
    \includegraphics[width=0.5\textwidth]{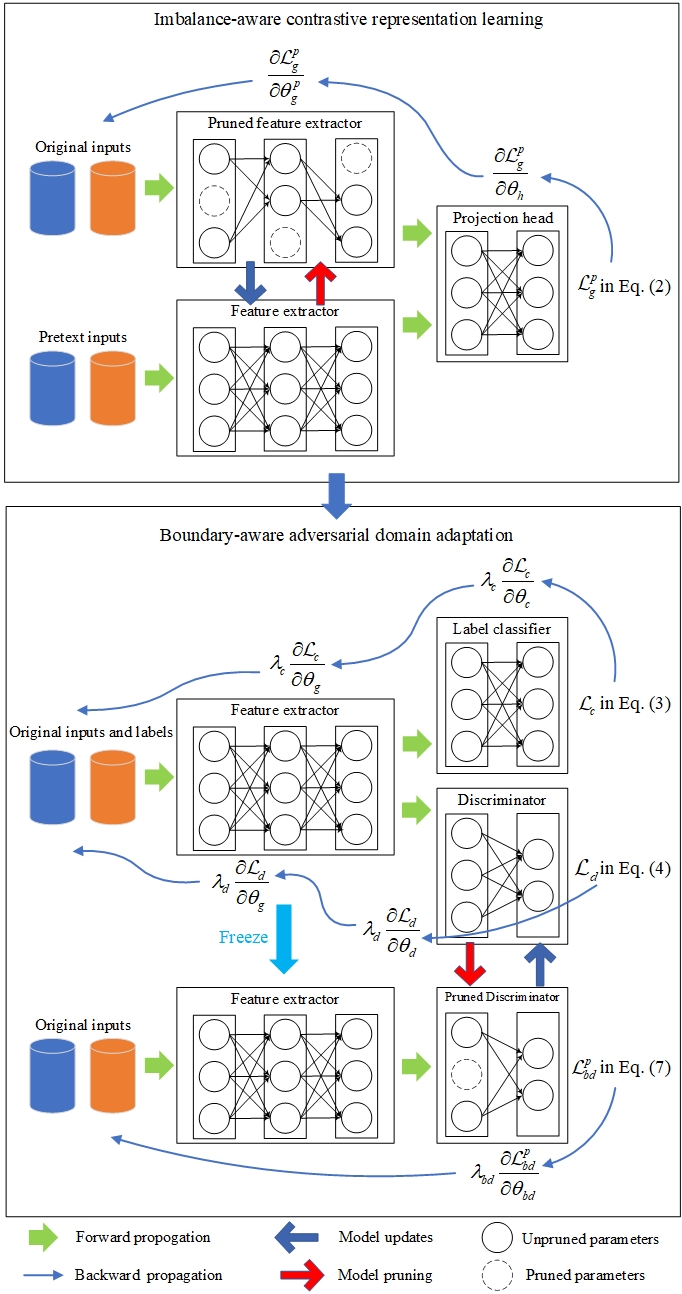}
    \caption{The overall structure of Sd-CDA. It consists of two parts: imbalance-aware contrastive representation learning and boundary-aware adversarial domain adaptation.}
    \label{fig:structure}
\end{figure}

Assume that the source domain $D_s=\{(X^i_s,y^i_s)\}_{i=1}^{n_s}$ and the target domain as $D_t=\{(X^i_t)\}_{i=1}^{n_t}$, where $n_s$ and $n_t$ are the sample size of the source and the target domain, respectively, $y^i_s \in \{1,..., K\}$ represents the classification label. Besides, these two domains have different marginal distributions, i.e., $P(X_s) \not= P(X_t)$, and different conditional distributions, i.e.,  $P_s(y|X) \not= P_t(y|X)$. The main goal is to train a network for the target task of cross-domain fault diagnosis, i.e., find accurate labels $y^i_t$'s for the target samples. Figure \ref{fig:structure} shows the overall structure of the proposed method. It consists of two parts: imbalance-aware contrastive representation learning (Ia-CLR) and boundary-aware adversarial domain adaptation (Ba-ADA). 

First, it implements contrastive learning to pre-train the feature extractor. The original inputs, including $X_s's$ and $X_t's$, and the corresponding augmented samples (pretext samples) constitute a new dataset for contrastive learning so that the pre-trained feature extractor can guarantee that its generated features are invariant to different domains. In order to relieve the possible influence of bi-imbalance during the iteration, we construct a new feature extractor by pruning the original one and formulate a pruned contrastive learning to pre-train the feature extractor. 

Then, it implements the boundary-aware adversarial domain adaptation to train an efficient classifier for the target domain. In the Ba-ADA, we first train the feature extractor, domain discriminator and label classifier together via domain adversarial training. Then, it freezes the feature extractor and label classifier and implements pruned supervised contrastive learning for the domain discriminator to detect the samples along the domain boundary. By repeating these steps, Ba-ADA can force the feature extractor to generate informative features rather than features along the domain boundary and thus train a more effective label classifier.

\subsection{Imbalance-aware contrastive learning of representations}
In this work, we only implement one augmentation to the inputs and regard each input with the augmented version as the positive pair. If we denote $X = X_s \cap X_t$ and the augmented version of $X$ as $\hat{X}$, then $(x_i, \hat{x}_i)$ is consider as a positive pair, where $i=1,...,n_s + n_t$. Let's denote the feature extractor as $G$ and the projection head as $P$. Original SimCLR feeds the positive or negative pairs to the same $G$ and $P$ and calculates the corresponding NT-Xent loss. In order to focus on feature learning for minorities, we create a pruned version of $G$ based on usual pruning tricks for neural networks \cite{molchanov2016pruning}, denoted as $G^p$. The pruning process mainly removes the weights or biases with small magnitudes or sets a small portion of the weights to zero randomly. The top part of Figure \ref{fig:structure} shows the detailed training process of the Ia-CLR. We first feed the pair separately $(x_i, \hat{x}_i)$ into two feature extractors, i.e., $G$ and $G^p$, and get a pair of generated features $(G(x_i), G^p(\hat{x}_i))$. Then they get projected as $(P(G(x_i)), P(G^p(\hat{x}_i)))$ through the same projection head $P$. We define the \textit{Pruned Normalized Temperature-Scaled Cross-Entropy loss} (PNT-Xent) to enforce the similarity between the positive pair $(P(G(x_i)), P(G^p(\hat{x}_i)))$. If we define $v_i=P(G(x_i))$ and $\hat{v}_i^+=P(G^p(\hat{x}_i))$ for simplicity, then we can construct the PNT-Xent of the whole system as follows:
\begin{equation}\label{pntloss}
    \mathcal{L}_g^p=\frac{1}{n_s+n_t}\sum_{i=1}^{n_s+n_t}-\log \frac{d(v_i,\hat{v}_i^+)}{d(v_i,\hat{v}_i^+) + \sum_{v_i^-\in V^-}d(v_i,v_i^-)},  
\end{equation}
where $V^- = \{(X \backslash v_i) \cup (\hat{X}\backslash \hat{v}_i^+\})\}$ represents the set of negative pairs. Based on \cite{hooker2019compressed}, if $x_i$ belongs to the minority, the difference between the corresponding projections $v_i$ and $\hat{v}_i^+$ should be larger than that of majority samples. As a result, the proportion of the loss caused by minorities should be more significant than that of majorities so that the training will be forced to pay more attention to minorities automatically. 

Compared with the original SimCLR, the Ia-CLR can detect minorities and enlarge their proportion in the loss function automatically to avoid bias toward majorities in the training process. Compared with the existing imbalance learning approaches in supervised settings, there is no need to search for optimal weights or pseudo-labels for the minorities to do re-weighting or re-balancing in Ia-CLR. More importantly, Ia-CLR is highly suitable for the UDA setting because it does not need the labels for the target samples.

\subsection{Boundary-aware adversarial domain adaptation}
The original domain adversarial learning may fail because the feature extractor can simply generate samples along the domain boundary to fool the domain discriminator, especially for the target samples, because we do not have the corresponding task-specific information available for target samples. To solve this problem, we propose the boundary-aware adversarial domain adaptation (Ba-ADA) framework to detect the features along the domain boundary and force the feature extractor to generate discriminative features for all samples. The main intuition of the Ba-ADA also comes from \cite{hooker2019compressed}, i.e., difficult samples are hard for DNNs to memorize well and may be easily forgotten them the DNNs are pruned. The samples along the domain boundary are difficult samples for $D$ with no doubt because the predictive results of these samples can change easily with even a small disturbance of $D$. 

Ba-ADA includes four parts: feature extractor $G$, domain discriminator $D$, pruned domain discriminator $D^p$ and the label classifier $C$. The bottom part of Figure \ref{fig:structure} shows the detailed structure of Ba-ADA. The training of Ba-ADA consists of three steps in each iteration: supervised learning, domain adversarial training, and boundary-aware contrastive learning.

In the supervised learning stage, we use the labeled source samples to train the feature extractor $G$ and label classifier $C$ to accurately predict labels for the source domain based on the label classification loss $\mathcal{L}_l$ defined as follows:
\begin{equation}
    \mathcal{L}_c =  -\frac{1}{n_s}\sum_{i=1}^{n_s}\sum_{k=1}^K \mathbf{1}_{[y_s = k]} \log(C(G(x_s^i))).
\end{equation}

Next, we implement adversarial training to narrow the discrepancy between the source domain and the target domain. The adversarial loss is defined as follows:
\begin{equation}
    \mathcal{L}_d = -\sum_{i=1}^{n_s}\log (D(G(x_s^i))) - \sum_{j=1}^{n_t}\log (1- D(G(x_t^j))). 
\end{equation}


Last, we introduce the boundary-aware constrastive learning. As discussed before, original adversarial training cannot detect the samples along the domain boundary, especially for the target samples, and thus the generated domain invariant features are not reliable. Supervised contrastive learning (SupCon) \cite{khosla2020supervised} is proposed to guide the model to output similar features if the inputs come from the same class. As a result, the SupCon has the potential to guide the training of $D$ to push the samples along the boundary back to the corresponding domain and make the output of $D$ given by each domain more compact. Following this idea, we first define the supervised contrastive domain adversarial loss (SupCon-DA). Denote the output of $D$ for $X_s$ and $X_t$ as $U_s = D(G(X_s))$ and $U_t = D(G(X_t))$ for simplicity. Also denote $U=U_s \cup U_t$ as the overall output space of $D$. Different from SimCLR, here we consider all the outputs in $U_s$ as positive pairs mutually and the same for $U_t$ to utilize the domain label information. By doing this, the SupCon-DA loss encourages $D$ to give close outputs to all features from the same domain and results in a more robust clustering of output space to avoid being misled by the samples along the domain boundary generated by $G$. Let's use one sample $u_s^i$ in $U_s$ to illustrate the definition of SupCon-DA loss:
\begin{equation}\label{samplesup}
    \mathcal{L}_{bd}(u_{s}^i) = \frac{1}{n_{s}-1}\sum_{q\in A_s(u_{s}^i)}\log \frac{d(u_{s}^i,q)}{\sum_{p \in B(u_{s}^i)}d(u_{s}^i,p)},
\end{equation}
where $A_s(u_s^i) = \{U_s \backslash u_s^i\}$ represents all the positive samples corresponding to $u_s^i$, and $B(u_{s}^i) = \{U \backslash u_{s}^i\}$ represents all the other samples in the overall space $U$. Please note that the loss \ref{samplesup} for the target samples also takes the same format by switching the subscripts $s$ to $t$ only. Then we can define the overall SupCon-DA loss of $U$ as follows:
\begin{equation}
    \mathcal{L}_{bd} = \frac{1}{n_s}\sum_{i=1}^{n_s}\mathcal{L}_{bd}(u_{s}^i) + \frac{1}{n_t}\sum_{i=1}^{n_t}\mathcal{L}_{bd}(u_{t}^i).
\end{equation}
If we fine-tune the domain discriminator $D$ by minimizing $\mathcal{L}_{bd}$, we can keep the predictive results of $D$ of different domains away from the boundary effectively.

However, this SupCon-DA learning is also vulnerable to the possible bi-imbalance just like what we discussed about SimCLR before. As the training process goes on, the difficult samples of the majority of the two domains can be well detected and pushed away from the boundary by $D$. However, the difficult samples of the minority can be easily neglected because the corresponding gradients may be diluted by the majority in the SupCon-DA loss. As a result, we construct a pruned version of the latest $D$, denoted as $D^p$, to help detect the difficult samples of the minority one step further. The intuition behind this approach is quite simple: given a difficult sample for the minority, the predictive domain labels of $D^p$ and $D$ can be absolutely different, thus intriguing a large disturbance to the corresponding SupCon-DA loss to draw attention from $D$. Here we use $U_s^p = D^p(G(X_s))$, $U_t^p = D^p(G(X_t^p))$ and $U^p = U^p_s \cup U_t^p$ to denote the outputs of $D^p$ correspondingly. Then we can define the \textit{Pruned Supervised Contrastive Domain Adversarial Loss} (PSupCon-DA) as:
\begin{equation}
    \mathcal{L}_{bd}^p = \frac{1}{n_s}\sum_{i=1}^{n_s}\mathcal{L}_{bd}(u_{sp}^i) + \frac{1}{n_t}\sum_{i=1}^{n_t}\mathcal{L}_{bd}(u_{tp}^i),
\end{equation} 
where $u_{sp}^i = D^p(G(x_s^i))$ and $u_{tp}^i = D^p(G(x_t^i))$. Please note that the original 

We fine-tune $D$ based on $\mathcal{L}_{cd}^p$ rather than $\mathcal{L}_{cd}$ to solve the imbalance challenge. 

\subsection{Training objectives and training procedures}
In this section, we will discuss the detailed training procedures and objectives of the proposed method. Let's denote the parameters of feature extractor $G$, pruned feature extractor $G^p$, projection head $H$, domain discriminator $D$, pruned domain discriminator $D^p$ and label classifier $C$ as $\theta_g$, $\theta_g^p$, $\theta_h$, $\theta_d$, $\theta_d^p$ and $\theta_c$, respectively. The ultimate objective of the proposed method can be written as:
\begin{equation}\label{originalobj}
    \min_{\theta_g, \theta_c}\max_{\theta_d} \mathcal{L}_g^p + \lambda_c\mathcal{L}_c - \lambda_d \mathcal{L}_d + \lambda_{bd}\mathcal{L}_{bd}^p,
\end{equation}
where $\lambda_c$, $\lambda_d$ and $\lambda_{bd}$ represent the weights of different loss, respectively. Next, we will introduce the training process of \ref{originalobj} in detail.

In the Ia-CLR, we train the feature extractor $G$ in a self-supervised manner to learn the feature representation of the source and the target samples, i.e., find the optimal $\theta_g$ and $\theta_h$ to minimize the PNT-Xent loss $\mathcal{L}_g^p$. Since $\mathcal{L}_g^p$ is calculated based on $G^p$ with parameter $\theta_g^p$ we cannot backpropagate the gradient of $\theta_g$ directly. As a result, we update the parameters in $\theta_g$ corresponding to $\theta_g^p$ and freeze the remaining parts (pruned parameters) in each iteration. Specifically, if we denote the $\theta_g = \{\theta_g^p, \bar{\theta}_g^p\}$ with $\bar{\theta}_g^p$ denoting the pruned parameters, the training process of IaCLR can be written as:
$$\theta_g \xrightarrow{p} \{\theta_g^p, \bar{\theta}_g^p\}$$
$$\theta_g^p \xleftarrow{} \theta_g^p -\omega\frac{\partial{\mathcal{L}_g^p} }{\partial \theta_g^p}$$
$$\theta_h \xleftarrow{} \theta_h -\omega\frac{\partial{\mathcal{L}_g^p} }{\partial \theta_h}$$
$$\theta_g \xleftarrow{} \{\theta_g^p, \bar{\theta}_g^p\}$$
where $\xrightarrow{p}$ represents pruning process and $\omega$ is the basic learning rate.

In Ba-ADA, we will use $\theta_g$ trained by IaCLR as the initial parameters of $G$. The training process of Ba-ADA can be divided into two parts. In the first part, we implement the original domain adversarial learning to train $G$, $D$ and $C$ as follows:
$$\theta_g \xleftarrow{} \theta_g - \lambda_c\omega\frac{\partial{\mathcal{L}_c} }{\partial \theta_g} + \lambda_d\omega \frac{\partial{\mathcal{L}_d} }{\partial \theta_g}$$
$$\theta_d \xleftarrow{} \theta_d - \lambda_d\omega \frac{\partial{\mathcal{L}_d} }{\partial \theta_d}$$
$$\theta_c \xleftarrow{} \theta_c - \lambda_c\omega\frac{\partial{\mathcal{L}_c} }{\partial \theta_c}$$

Next, we use the PSupCon-DA loss to fine-tune $D$ based on the pruned version $D^p$. Similar to the IaCLR, we cannot update $D$ by minimizing $\mathcal{L}_{bd}^p$ directly. Again, we denote $\theta_d = \{\theta_d^p, \bar{\theta}_d^p\}$ then the training process of SupCon-DA can be written as:
$$\theta_d \xrightarrow{p} \{\theta_d^p, \bar{\theta}_d^p\}$$
$$\theta_d^p \xleftarrow{} \theta_d^p -\lambda_{bd}\omega \frac{\partial{\mathcal{L}_{bd}^p} }{\partial \theta_d^p}$$
$$\theta_d \xleftarrow{} \{\theta_d^p, \bar{\theta}_d^p\}$$

\subsection{Implementation details}
In this section, we will discuss some guidelines for the choice of hyper-parameters and model structure in the proposed method.
\subsubsection{Structure of neural networks}
In Figure \ref{fig:structure}, we only use some simple structures of fully connected networks (FNNs) to illustrate the proposed method. However, in real-world applications, we need to design different structures applicable to data with different properties. For example, we need to incorporate the CNN-pool module when facing vibration signals or images but use FNNs or Resnet structures for structured data. We will show different structures in our experiments. 
\subsubsection{Pruning methods}
The prevailing pruning methods include random pruning methods \cite{han2015learning} and magnitude-based pruning methods \cite{han2015deep}. Random pruning methods set a proportion of a model's parameters to zero randomly while magnitude-based pruning methods mask the model's parameters with the smallest absolute values based on a predefined proportion hyper-parameter. To maintain a stable training process and avoid large vibration of the pruned model-related loss, we implement magnitude-based pruning methods based on the L1 norm \cite{molchanov2016pruning}. As for the choice of proportion hyper-parameter $\alpha_g$ in IaCLR and $\alpha_d$ in Ba-ADA, we have several empirical guidelines based on our experiments. 

In IaCLR, a higher $\alpha_g$ often leads to a faster decrease of $\mathcal{L}_g^p$ at the beginning stage because a higher $\alpha_g$ causes a larger disagreement between $G$ and $G^p$. However, the loss may not decrease after a certain amount of iterations or even increase as the training process, which we call the "bottleneck" of training. This bottleneck is due to the large difference between the feature generator and its pruned model. We cannot guarantee converged training when the bottleneck appears. As a result, we recommend a relatively small $\alpha_g$, i.e., $0.1<\alpha_g<0.3$ for Ia-CLR to maintain stable training. 

In Ba-ADA, we empirically find that a larger $\alpha_d$ generally leads to a higher accuracy for the the final classifier. The possible reason is due to the motivation of PSupCon-DA, i.e., higher $\alpha_d$ leads to a more significant disturbance of $\mathcal{L}_{bd}^p$ to remind it of more difficult samples. However, a larger disturbance of $\mathcal{L}_{bd}^p$ may cause gradient explosion and make the whole training diverge. As a result, we recommend choosing the optimal $\alpha$ for PSupCon-DA carefully based on the corresponding loss value of $\mathcal{L}_{bd}^p$. Based on our experiments, we would recommend $0.3<\alpha_d<0.5$ for SupCon-DA. We will show detailed experiments regarding the choice of $0.3<\alpha_d<0.5$ in Section \ref{experiments}.

\subsubsection{Basic learning rate $\omega$ and loss weights $\lambda_c$, $\lambda_d$ and $\lambda_{bd}$}
Different from the training of traditional neural networks, domain adaptation-related methods cannot use accuracy on the validation set to set the learning rate and stop criterion because we do not have labels for the target domain. In this work, we choose the basic learning rate $\omega$ according to the loss trend of the Ia-CLR training process. Since a higher $\omega$ also means a faster decrease of $\mathcal{L}_g^p$ and may get bottleneck just like what a higher $\alpha_g$ may cause in Ia-CLR. As a result, we will recommend a small $\omega$ as long as the training process can proceed and the overall training process does not exceed the time requirements.

Loss weights $\lambda_c$ and $\lambda_d$ determine the adversarial training of $G$ and $D$. Generally, $G$ has a more complex structure than $D$ which means $\theta_g$ needs more training than $\theta_d$. As a result, we recommend setting $\lambda_c > \lambda_d$. As for $\lambda_{bd}$, we also recommend a relatively small value to guarantee stable training, i.e., $\lambda_{bd}<\lambda_d$.





\section{Case studies}\label{experiments}
In this section, we will evaluate the proposed Sd-CDA on two different cases, i.e., a mechanical rolling bearing dataset and an industrial three-phase flow process dataset. We compare the proposed method with five different approaches including the original CNNs, joint domain adaptation (JDA) \cite{long2014transfer}, domain adversarial neural networks (DANN) \cite{ganin2016domain}, contrastive domain adaptation (ConDA) \cite{azuma2023adversarial} and an imbalanced domain adaptation (Imba-DA) \cite{wu2022imbalanced}.

\subsection{Mechanical rolling bearing case}
The mechanical rolling bearing dataset provided by Case Western Reserve University includes vibration signals collected from different positions of the bearing \cite{smith2015rolling}, i.e., fan end and drive end, and different working speeds, i.e., 1730 rpm, 1750 rpm, 1772 rpm and 1797 rpm. In this experiment, we choose the data collected on the drive end with a sampling frequency of 12kHz and consider two working speeds with the largest differences, i.e., 1730 rpm and 1797 rpm, to construct two different domain adaptation cases, i.e., 1730 rpm $\xrightarrow{}$ 1797 rpm and 1797 rpm $\xrightarrow{}$ 1730 rpm.  

\begin{figure*}[ht]
    \centering
    \includegraphics[width=\textwidth]{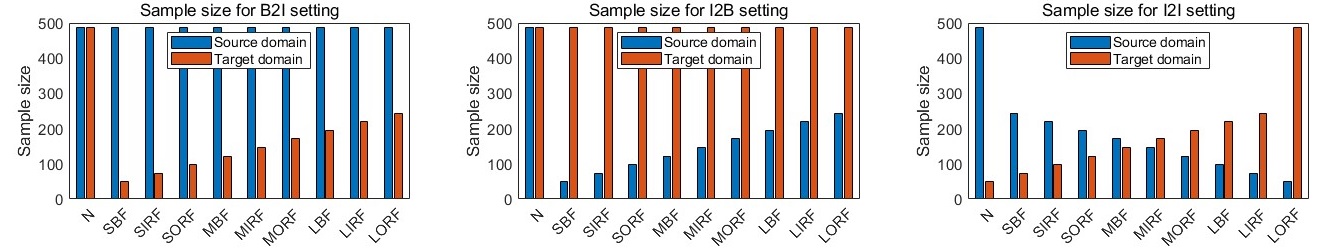}
    \caption{Sample size for CWRU dataset in different settings}
    \label{fig:CWRUsize}
\end{figure*}

Specifically, this dataset consists of a normal condition (N) and a total of nine fault categories: three different fault modes, i.e., inner ring fault (IR), ball fault (BF) and outer ring fault (ORF), and three different fault diameters, i.e., 0.007 inch (S), 0.014 inch (M) and 0.028 inch (L). We construct one balance setting and three imbalance settings to simulate the possible bi-imbalance to test the performance of the proposed domain adaptation method, i.e., balanced source domain to balanced target domain (B2B), balanced source domain to imbalanced target domain (B2I), imbalanced source domain to balanced target domain (I2B), imbalanced source domain to imbalanced target domain (I2I). As a result, we have a total of eight scenarios as shown in Table \ref{tab:settings}. Since the original dataset is approximately balanced, we use all the original samples of two domains to formulate the B2B setting. As for the other three imbalanced settings, we sample the original dataset for different categories based on the sample size as shown in Figure \ref{fig:CWRUsize}. As we can see, the B2I and I2B settings simulate the scenario with bi-imbalance, and the bi-imbalance is much more severe in the I2I setting setting.
    \begin{table}[ht]
        \centering
        \caption{All scenarios with different cases and different imbalance settings.}
        \begin{tabular}{cccc}
        \toprule
             Case & Source & Target & Imbalance settings \\
        
        \midrule
              1 & 1797 rpm  & 1730 rpm & B2B, B2I, I2B, I2I \\
        \midrule
              2 & 1730 rpm & 1797 rpm & B2B, B2I, I2B, I2I  \\
        \bottomrule
        \end{tabular}
        \label{tab:settings}
    \end{table}
    
In order to extract the information in the vibration signals, we follow \cite{shao2018highly} to convert the original data into $18\times 18$ images by wavelet transform. A CNN-based structure with Conv($3\times3$, 32)-Pool-Conv($3\times3$, 64)-Pool-Conv($3\times3$, 128) is used as the basic feature extractor to construct different diagnosis structures where Conv($3\times3$, 128) means that the kernel size is $3\times3$ and the number of output channels is 128. The basic domain discriminator, basic projection head and basic label classifier are all set to FNNs with two layers and the basic learning rate $\omega=0.001$. The implementation details of the six methods are as follows:
\begin{itemize}
    \item CNN: Basic feature extractor + basic label classifier;
    \item JDA: Basic feature extractor + basic label classifier, regularize parameter $\lambda = 1$;
    \item DANN: Basic feature extractor + basic label classifier and basic domain discriminator;
    \item ConDANN: basic feature extractor + basic projection head as the SimCLR model + DANN;
    \item Imba-DA: Basic feature extractor + basic label classifier and basic domain discriminator with $\mu_1 = 0.95$ and $\mu_2 = 0.05$;
    \item Sd-CDA: Basic feature extractor + basic projection head as the Ia-CLR mode where the pruning method is L1-norm pruning with pruning proportion $\alpha_g=0.25$. DANN with $\lambda_c=\lambda_d = 1$ and $\lambda_{bd} = 10^{-7}$ as the Ba-ADA model and the L1-norm pruning with $\alpha=0.4$ is used to prune the domain discriminator. 
\end{itemize}

We average the diagnosis accuracy of two domain adaptation cases and list them in Table \ref{tab:CWRUacc}. The classic DA approaches JDA and DANN reach good accuracy for the B2B setting but their accuracy drops dramatically for the three imbalanced settings because they do not consider the imbalance. With the help of contrastive learning, the ConDA gets higher accuracy compared with DANN because the pre-tain process helps ConDA learn the feature representation more efficiently, but it still fails for imbalanced settings. On the basis of DANN, the imba-DA gets higher accuracy because it considers the imbalance distributions of two domains and implements cost-sensitive re-weighting to guide the training. However, the cost-sensitive policy cannot detect the samples on the domain boundary, especially for the target samples. The proposed Sd-CDA beats other DA or imbalanced DA approaches for three imbalanced settings, especially for the I2B and I2I settings.

\begin{figure*}[ht]
    \centering
    \includegraphics[width=\textwidth]{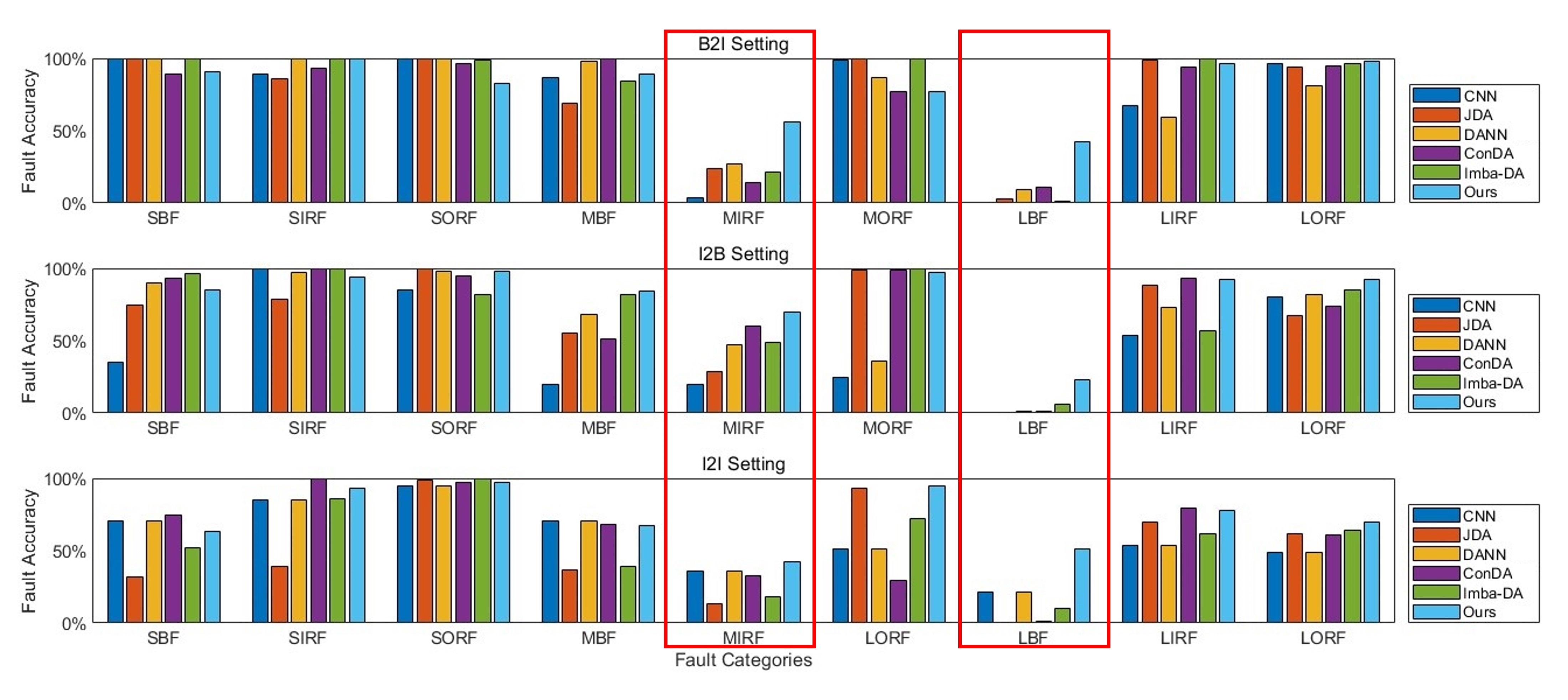}
    \caption{Class-wise diagnosis accuracy for CWRU dataset.}
    \label{fig:CWRUacc}
\end{figure*}

Figure \ref{fig:CWRUacc} shows the detailed diagnosis accuracy for all categories under three imbalanced settings. We can see that the MIRF and LBF faults are difficult to detect in general. Compared with other approaches, the proposed method can get approximately the same performance for other 'easy' categories, but for the 'difficult' categories, the proposed method performs much better than others. Especially for the LBF fault, the proposed method achieves at least twice the recognition accuracy compared to other methods. Figure \ref{fig:CWRUacc} proves the superiority of Ia-CLR and Ba-ADA towards the imbalanced samples.

    \begin{table}[ht]
        \centering
        \caption{Average diagnosis accuracy for different methods under four settings.}
        \begin{tabular}{ccccc}
        \toprule
             Methods & B2B & B2I & I2B & I2I  \\
        
        \midrule
             CNN & 82.43 & 80.44 & 62.51 & 51.19  \\
        \midrule
             JDA  & 92.34 & 84.24 & 72.01 & 54.49  \\
        \midrule
             DANN & 87.27 & 83.92 & 72.31 & 56.04   \\
        \midrule
             ConDA & 92.26 & 85.64 & 78.24 & 58.24  \\
        \midrule
             Imba-DA & 91.37 & 86.42 & 79.23 & 64.58  \\
        \midrule
             Ours & 91.26 & 90.21 & 84.92 & 69.54  \\
        \bottomrule
        \end{tabular}
        \label{tab:CWRUacc}
    \end{table}

\subsection{Industrial three-phase flow process}
\begin{figure}[ht]
    \centering
    \includegraphics[width=0.5\textwidth]{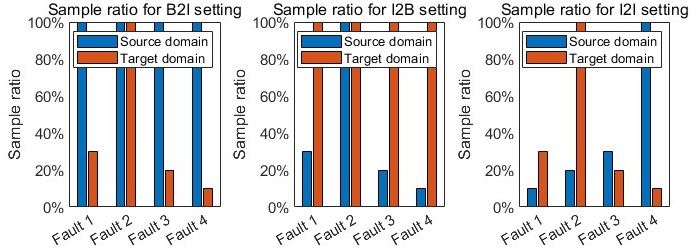}
    \caption{Sample ratio for Three-phase flow process in imbalanced settings}
    \label{fig:Triplesize}
\end{figure}

Cranfield University's three-phase flow facility is designed to deliver precise, controlled flow rates of water, oil, and air into a pressurized system \cite{ruiz2015statistical}. The facility can manage either single phases of these fluids or their combinations in specific proportions. During this experimentation, two primary process inputs, i.e., the air and water flow rates, and the other 24 variables are considered. All data is collected with a 1 Hz frequency. Notably, the facility allows for the selection of four distinct air flow rates and five water flow rates, which leads to different working conditions and data distributions. In this experiment, we choose the data under two different water flow rate as our domain A and domain B and consider four faults. The detailed information of the original datasets is shown in Table \ref{tab:Triplesettings}. We also follow Table \ref{tab:settings} to construct two domain adaptation cases based on domain A and domain B and test the performance of different methods under B2B, B2I, I2B, and I2I settings. Figure \ref{fig:Triplesize} shows the sample ratio on the basis of the original dataset in different imbalanced settings.

\begin{table}[htbp]
    \centering
    \caption{Dataset information of three-phase flow process}
    \begin{tabular}{@{}cccccc@{}}
        \toprule
        & \multicolumn{2}{c}{Domain A} & \multicolumn{2}{c}{Domain B} \\
        \cmidrule(lr){2-3} \cmidrule(lr){4-5}
        Fault class & Water flow  & No. of & Water flow  & No. of \\
        & rate  & samples   & rate  & samples \\
        \midrule
        Fault 1 & 2 & 3121 & 3.5 & 3001 \\
        Fault 2 & 2 & 2181 & 3.5 & 2137 \\
        Fault 3 & 2 & 3000 & 3.5 & 3001 \\
        Fault 4 & 2 & 3000 & 3.5 & 3001 \\
        \bottomrule
    \end{tabular}
    \label{tab:Triplesettings}
\end{table}

In this experiment, we use the FNNs to construct the feature extractor. Specifically, an FNN(100)-Dropout(0.3)-FNN(24)-Dropout(0.3) structure is used as the basic feature extractor. The basic domain discriminator, basic projection head and basic label classifier are all set to FNNs with two layers and the basic learning rate $\omega=0.0001$. The implementation details of different methods are as follows:
\begin{itemize}
    \item FNN: Basic feature extractor + basic label classifier;
    \item JDA: Basic feature extractor + basic label classifier, regularize parameter $\lambda = 1$;
    \item DANN: Basic feature extractor + basic label classifier and basic domain discriminator;
    \item ConDANN: basic feature extractor + basic projection head as the SimCLR model + DANN;
    \item Imba-DA: Basic feature extractor + basic label classifier and basic domain discriminator with $\mu_1 = 0.95$ and $\mu_2 = 0.05$;
    \item Sd-CDA: Basic feature extractor + basic projection head as the Ia-CLR mode where the pruning method is L1-norm pruning with pruning proportion $\alpha=0.25$. DANN with $\lambda_c=1$, $\lambda_d = 0.001$ and $\lambda_{bd} = 10^{-8}$ as the Ba-ADA model and the L1-norm pruning with $\alpha=0.3$ is used to prune the domain discriminator. 
\end{itemize}

Table \ref{tab:Tripleacc} lists the average diagnosis accuracy of two domain adaptation cases, i.e., A to B and B to A. We can see that the proposed method achieves the best performance in general, especially for the B2I and I2I settings. One can see that in the I2I setting, the classific DA approaches almost fail because the impact of imbalance hinders the classifier heavily. The proposed method gets much higher accuracy, over twice the accuracy compared with the original DANN, which proves that the proposed method is able to relieve the bi-imbalance and get a satisfactory classifier. 

    \begin{table}[ht]
        \centering
        \caption{Average diagnosis accuracy for different methods under four settings}
        \begin{tabular}{ccccc}
        \toprule
             Methods & B2B & B2I & I2B & I2I \\
        
        \midrule
             FNN & 65.54 & 42.12 & 55.51 & 21.21  \\
        \midrule
             JDA  & 72.32 & 40.22 & 54.01 & 22.32  \\
        \midrule
             DANN & 71.32 & 65.54 & 43.53 & 21.24  \\
        \midrule
             ConDA & 56.21 & 59.42 & 46.25 & 44.26  \\
        \midrule
             Imba-DA & 78.54 & 62.32 & 61.45 & 55.23 \\
        \midrule
             Ours & 84.23 & 66.23 & 61.25 & 58.25 \\
        \bottomrule
        \end{tabular}
        \label{tab:Tripleacc}
    \end{table}

\begin{figure*}[ht]
    \centering
    \includegraphics[width=\textwidth]{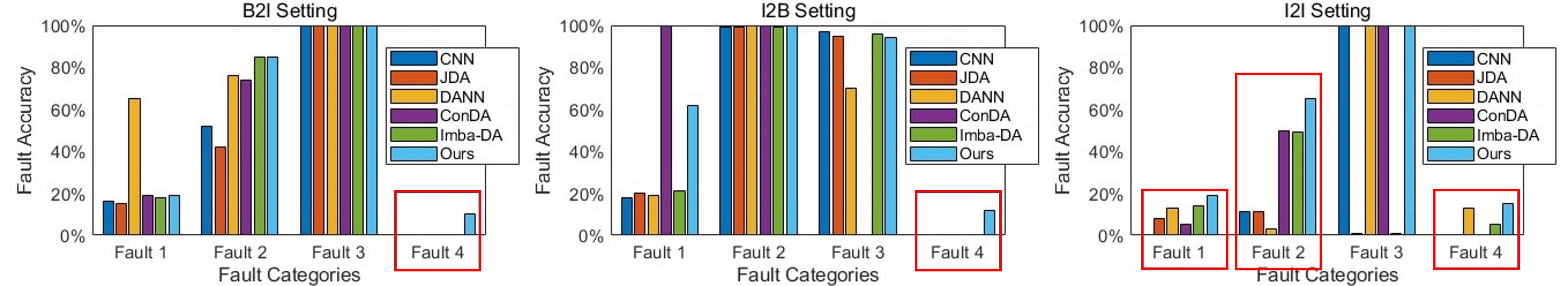}
    \caption{Class-wise diagnosis accuracy for three-phase flow process data.}
    \label{fig:Tripleacc}
\end{figure*}

Figure \ref{fig:Tripleacc} shows the class-wise accuracy for three imbalanced settings. We can see that almost all other methods cannot detect fault 4 but the proposed method can achieve a certain amount of detection. As for the other three faults in B2I and I2B settings, the proposed method gets approximately the same good performance as other methods. In the B2I setting, DANN gets higher accuracy for fault 1 but lower accuracy for fault 2 and fault 4 compared with the proposed method, so the proposed method still gets better overall accuracy. In the I2B setting, even though ConDA detects fault 1 better, it fails for fault 3 and fault 4. In the I2I setting, the superiority of the proposed method is quite clear for all faults.

\subsection{In-depth studies of the proposed method}
\subsubsection{Analysis for different $\alpha_d$ and $\lambda_{bd}$ for the proposed method}
In this experiment, we test the performance of the proposed method under different $\alpha_d$ and $\lambda_{bd}$ to explore some empirical guidelines. We use the 1797 $\xrightarrow{}$ 1730 setting to implement the experiment.

\begin{figure}[ht]
    \centering
    \includegraphics[width=0.5\textwidth]{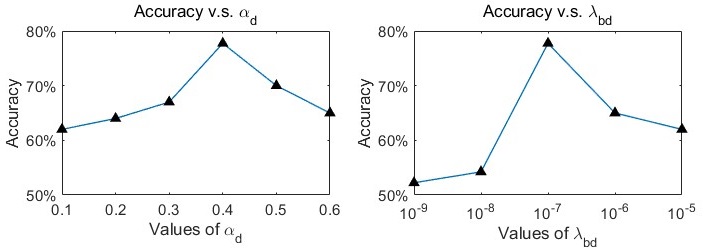}
    \caption{Accuracy of the proposed method under different $\alpha_d$ and $\lambda_{bd}$}
    \label{fig:different}
\end{figure}

First, we test the proposed method using different pruning proportions $\alpha_d$ for the Ba-ADA, ranging from 0.1 to 0.6. In this experiment, we use L1 pruning methods for Ia-CLR and Ba-ADA and set $\alpha_g=0.25$,  $\lambda_c=\lambda_d = 1$ and $\lambda_{bd} = 10^{-7}$. The left subfigure of Figure \ref{fig:different} shows the accuracy under different $\alpha_d$. When $\alpha_d$ is small, the accuracy increases as $\alpha_d$. The possible reason is that a higher $\alpha_d$ will cause larger $\mathcal{L}_{bd}^p$ leading to a heavier push to the discriminator to detect and pay attention to the sample along the boundary. However, if the resulted $\mathcal{L}_{bd}^p$ is too large, it may cover the gradient of $\mathcal{L}_{d}$ and $\mathcal{L}_c$. As a result, we cannot reach efficient domain adaptation. More importantly, the $\mathcal{L}_{bd}^p$ will remain large and cannot decrease so that the overall training process will diverge if a large $\alpha_d$ is used. As a result,  the accuracy drops dramatically when $\alpha_d$ increases after 0.4.  

Then we test the proposed method using different $\lambda_{bd}$, ranging from $10^{-9}$ to $10^{-5}$. In this experiment, we use L1 pruning methods for Ia-CLR and Ba-ADA and set $\alpha_g=0.25$, $\alpha_d=0.4$  $\lambda_c=\lambda_d = 1$. The left subfigure of Figure \ref{fig:different} shows the accuracy of this experiment. We can see that the $\lambda_{bd}$ also shows the same trend as the $\alpha_d$ because a larger $\lambda_{bd}$ also leads to a larger update step caused by $\mathcal{L}_{bd}^p$. Different from the impact of different $\alpha_d$, the accuracy changes more heavily along $\lambda_{bd}$. As a result, we consider $\lambda_{bd}$ as the coarse adjustment and $\alpha_d$ as the fine adjustment of the proposed method. In other words, we need to first choose a suitable $\lambda_{bd}$ to make sure a stable changing process of $\mathcal{L}_{bd}^p$ and then tune the $\alpha_d$ to get better accuracy. 

\subsubsection{Ablation study}
We also implement ablation studies to test the efficiency of IaCLR and BaADA separately based on the 1797 $\xrightarrow{}$ 1730 setting. The benchmark is chosen as the original DANN. We add two key parts of the Sd-CDA step by step on the basis of DANN and generate three situations, i.e., DANN + Ia-CLR, DANN + Ba-ADA, and DANN + Ia-CLR + Ba-ADA (Sd-CDA) as shown in Table \ref{tab:ablation}.

From Table \ref{tab:ablation} we can see that, these two components are all important to the proposed Sd-CDA, i.e., they all contribute to a better diagnosis accuracy, especially in the three situations with imbalance. By comparison, the contribution of Ba-ADA tends to be slightly lower than Ia-CLR in B2I and I2B situations but higher in the I2I situation. This phenomenon is quite reasonable. In B2I and I2B situations, the most decisive question is to learn effective feature representation to leverage the information of samples because we have more samples. As a result, the pre-train strategy stands out. But in the I2I situation, the imbalances are quite severe and also different in the two domains. In this situation, how to handle the imbalance when training a stable classifier becomes more important. This is why BaADA plays a significant role in the I2I situation.

    \begin{table}[ht]
        \centering
        \caption{Ablation study for the proposed method}
        \begin{tabular}{cccccc}
        \toprule
             Ia-CLR & Ba-ADA & B2B & B2I & I2B & I2I \\
        
        \midrule
              &  & 89.27 & 86.24 & 74.58 & 64.78 \\
        \midrule
              $\checkmark$&  & 86.34 & 87.35 & 82.34 & 69.42  \\
        \midrule
              & $\checkmark$ & 88.25 & 86.25 & 80.05 & 74.25  \\
        \midrule
              $\checkmark$& $\checkmark$ & 93.14 & 91.14 & 86.30 & 77.74  \\

        \bottomrule
        \end{tabular}
        \label{tab:ablation}
    \end{table}

\section{Conclusion}\label{conclusion}
In this work, we propose an Sd-CDA framework for the industrial fault diagnosis tasks with both domain discrepancy and bi-imbalance challenges. It first pre-trains the feature extractor via pruned contrastive learning to learn efficient feature representations for the bi-imbalance data. Then it constructs a pruned supervised contrastive domain adversarial learning approach to avoid the learned features biased to the source domain or along the domain boundary. Additionally, we provide several guidelines for choosing the hyper-parameters of the proposed framework based on the empirical experiments. A series of experiments validate the superiority of our work.


\bibliography{ref}

\begin{thebibliography}{10}
\providecommand{\url}[1]{#1}
\csname url@samestyle\endcsname
\providecommand{\newblock}{\relax}
\providecommand{\bibinfo}[2]{#2}
\providecommand{\BIBentrySTDinterwordspacing}{\spaceskip=0pt\relax}
\providecommand{\BIBentryALTinterwordstretchfactor}{4}
\providecommand{\BIBentryALTinterwordspacing}{\spaceskip=\fontdimen2\font plus
\BIBentryALTinterwordstretchfactor\fontdimen3\font minus
  \fontdimen4\font\relax}
\providecommand{\BIBforeignlanguage}[2]{{%
\expandafter\ifx\csname l@#1\endcsname\relax
\typeout{** WARNING: IEEEtran.bst: No hyphenation pattern has been}%
\typeout{** loaded for the language `#1'. Using the pattern for}%
\typeout{** the default language instead.}%
\else
\language=\csname l@#1\endcsname
\fi
#2}}
\providecommand{\BIBdecl}{\relax}
\BIBdecl

\bibitem{yang2022paradigm}
Z.~Yang and Z.~Ge, ``On paradigm of industrial big data analytics: From
  evolution to revolution,'' \emph{IEEE Transactions on Industrial
  Informatics}, vol.~18, no.~12, pp. 8373--8388, 2022.

\bibitem{wu2018deep}
H.~Wu and J.~Zhao, ``Deep convolutional neural network model based chemical
  process fault diagnosis,'' \emph{Computers \& chemical engineering}, vol.
  115, pp. 185--197, 2018.

\bibitem{song2018fault}
B.~Song and H.~Shi, ``Fault detection and classification using
  quality-supervised double-layer method,'' \emph{IEEE Transactions on
  Industrial Electronics}, vol.~65, no.~10, pp. 8163--8172, 2018.

\bibitem{ding2023deep}
Y.~Ding, M.~Jia, J.~Zhuang, Y.~Cao, X.~Zhao, and C.-G. Lee, ``Deep imbalanced
  domain adaptation for transfer learning fault diagnosis of bearings under
  multiple working conditions,'' \emph{Reliability Engineering \& System
  Safety}, vol. 230, p. 108890, 2023.

\bibitem{li2020systematic}
C.~Li, S.~Zhang, Y.~Qin, and E.~Estupinan, ``A systematic review of deep
  transfer learning for machinery fault diagnosis,'' \emph{Neurocomputing},
  vol. 407, pp. 121--135, 2020.

\bibitem{chen2023transfer}
H.~Chen, H.~Luo, B.~Huang, B.~Jiang, and O.~Kaynak, ``Transfer
  learning-motivated intelligent fault diagnosis designs: A survey, insights,
  and perspectives,'' \emph{IEEE Transactions on Neural Networks and Learning
  Systems}, 2023.

\bibitem{gretton2012kernel}
A.~Gretton, K.~M. Borgwardt, M.~J. Rasch, B.~Sch{\"o}lkopf, and A.~Smola, ``A
  kernel two-sample test,'' \emph{The Journal of Machine Learning Research},
  vol.~13, no.~1, pp. 723--773, 2012.

\bibitem{sun2016deep}
B.~Sun and K.~Saenko, ``Deep coral: Correlation alignment for deep domain
  adaptation,'' in \emph{Computer Vision--ECCV 2016 Workshops: Amsterdam, The
  Netherlands, October 8-10 and 15-16, 2016, Proceedings, Part III 14}.\hskip
  1em plus 0.5em minus 0.4em\relax Springer, 2016, pp. 443--450.

\bibitem{ganin2016domain}
Y.~Ganin, E.~Ustinova, H.~Ajakan, P.~Germain, H.~Larochelle, F.~Laviolette,
  M.~March, and V.~Lempitsky, ``Domain-adversarial training of neural
  networks,'' \emph{Journal of machine learning research}, vol.~17, no.~59, pp.
  1--35, 2016.

\bibitem{chai2020fine}
Z.~Chai and C.~Zhao, ``A fine-grained adversarial network method for
  cross-domain industrial fault diagnosis,'' \emph{IEEE Transactions on
  Automation Science and Engineering}, vol.~17, no.~3, pp. 1432--1442, 2020.

\bibitem{saito2018maximum}
K.~Saito, K.~Watanabe, Y.~Ushiku, and T.~Harada, ``Maximum classifier
  discrepancy for unsupervised domain adaptation,'' in \emph{Proceedings of the
  IEEE conference on computer vision and pattern recognition}, 2018, pp.
  3723--3732.

\bibitem{noroozi2016unsupervised}
M.~Noroozi and P.~Favaro, ``Unsupervised learning of visual representations by
  solving jigsaw puzzles,'' in \emph{European conference on computer
  vision}.\hskip 1em plus 0.5em minus 0.4em\relax Springer, 2016, pp. 69--84.

\bibitem{bachman2019learning}
P.~Bachman, R.~D. Hjelm, and W.~Buchwalter, ``Learning representations by
  maximizing mutual information across views,'' \emph{Advances in neural
  information processing systems}, vol.~32, 2019.

\bibitem{chen2020simple}
T.~Chen, S.~Kornblith, M.~Norouzi, and G.~Hinton, ``A simple framework for
  contrastive learning of visual representations,'' in \emph{International
  conference on machine learning}.\hskip 1em plus 0.5em minus 0.4em\relax PMLR,
  2020, pp. 1597--1607.

\bibitem{lu2022domain}
N.~Lu, H.~Xiao, Z.~Ma, T.~Yan, and M.~Han, ``Domain adaptation with
  self-supervised learning and feature clustering for intelligent fault
  diagnosis,'' \emph{IEEE Transactions on Neural Networks and Learning
  Systems}, 2022.

\bibitem{azuma2023adversarial}
C.~Azuma, T.~Ito, and T.~Shimobaba, ``Adversarial domain adaptation using
  contrastive learning,'' \emph{Engineering Applications of Artificial
  Intelligence}, vol. 123, p. 106394, 2023.

\bibitem{jiang2021self}
Z.~Jiang, T.~Chen, B.~J. Mortazavi, and Z.~Wang, ``Self-damaging contrastive
  learning,'' in \emph{International Conference on Machine Learning}.\hskip 1em
  plus 0.5em minus 0.4em\relax PMLR, 2021, pp. 4927--4939.

\bibitem{hooker2019compressed}
S.~Hooker, A.~Courville, G.~Clark, Y.~Dauphin, and A.~Frome, ``What do
  compressed deep neural networks forget?'' \emph{arXiv preprint
  arXiv:1911.05248}, 2019.

\bibitem{lu2016deep}
W.~Lu, B.~Liang, Y.~Cheng, D.~Meng, J.~Yang, and T.~Zhang, ``Deep model based
  domain adaptation for fault diagnosis,'' \emph{IEEE Transactions on
  Industrial Electronics}, vol.~64, no.~3, pp. 2296--2305, 2016.

\bibitem{jiao2020residual}
J.~Jiao, M.~Zhao, J.~Lin, and K.~Liang, ``Residual joint adaptation adversarial
  network for intelligent transfer fault diagnosis,'' \emph{Mechanical Systems
  and Signal Processing}, vol. 145, p. 106962, 2020.

\bibitem{li2021central}
X.~Li, Y.~Hu, J.~Zheng, M.~Li, and W.~Ma, ``Central moment discrepancy based
  domain adaptation for intelligent bearing fault diagnosis,''
  \emph{Neurocomputing}, vol. 429, pp. 12--24, 2021.

\bibitem{chen2020domain}
Z.~Chen, G.~He, J.~Li, Y.~Liao, K.~Gryllias, and W.~Li, ``Domain adversarial
  transfer network for cross-domain fault diagnosis of rotary machinery,''
  \emph{IEEE Transactions on Instrumentation and Measurement}, vol.~69, no.~11,
  pp. 8702--8712, 2020.

\bibitem{huang2021multisource}
Z.~Huang, Z.~Lei, G.~Wen, X.~Huang, H.~Zhou, R.~Yan, and X.~Chen, ``A
  multisource dense adaptation adversarial network for fault diagnosis of
  machinery,'' \emph{IEEE Transactions on Industrial Electronics}, vol.~69,
  no.~6, pp. 6298--6307, 2021.

\bibitem{jiao2022self}
J.~Jiao, H.~Li, and J.~Lin, ``Self-training reinforced adversarial adaptation
  for machine fault diagnosis,'' \emph{IEEE Transactions on Industrial
  Electronics}, 2022.

\bibitem{zareapoor2021oversampling}
M.~Zareapoor, P.~Shamsolmoali, and J.~Yang, ``Oversampling adversarial network
  for class-imbalanced fault diagnosis,'' \emph{Mechanical Systems and Signal
  Processing}, vol. 149, p. 107175, 2021.

\bibitem{xiao2019transfer}
D.~Xiao, Y.~Huang, C.~Qin, Z.~Liu, Y.~Li, and C.~Liu, ``Transfer learning with
  convolutional neural networks for small sample size problem in machinery
  fault diagnosis,'' \emph{Proceedings of the Institution of Mechanical
  Engineers, Part C: Journal of Mechanical Engineering Science}, vol. 233,
  no.~14, pp. 5131--5143, 2019.

\bibitem{kuang2021class}
J.~Kuang, G.~Xu, T.~Tao, and Q.~Wu, ``Class-imbalance adversarial transfer
  learning network for cross-domain fault diagnosis with imbalanced data,''
  \emph{IEEE Transactions on Instrumentation and Measurement}, vol.~71, pp.
  1--11, 2021.

\bibitem{liu2022cross}
X.~Liu, J.~Chen, K.~Zhang, S.~Liu, S.~He, and Z.~Zhou, ``Cross-domain
  intelligent bearing fault diagnosis under class imbalanced samples via
  transfer residual network augmented with explicit weight self-assignment
  strategy based on meta data,'' \emph{Knowledge-Based Systems}, vol. 251, p.
  109272, 2022.

\bibitem{wu2022imbalanced}
Z.~Wu, H.~Zhang, J.~Guo, Y.~Ji, and M.~Pecht, ``Imbalanced bearing fault
  diagnosis under variant working conditions using cost-sensitive deep domain
  adaptation network,'' \emph{Expert Systems with Applications}, vol. 193, p.
  116459, 2022.

\bibitem{khosla2020supervised}
P.~Khosla, P.~Teterwak, C.~Wang, A.~Sarna, Y.~Tian, P.~Isola, A.~Maschinot,
  C.~Liu, and D.~Krishnan, ``Supervised contrastive learning,'' \emph{Advances
  in neural information processing systems}, vol.~33, pp. 18\,661--18\,673,
  2020.

\bibitem{molchanov2016pruning}
P.~Molchanov, S.~Tyree, T.~Karras, T.~Aila, and J.~Kautz, ``Pruning
  convolutional neural networks for resource efficient inference,'' \emph{arXiv
  preprint arXiv:1611.06440}, 2016.

\bibitem{han2015learning}
S.~Han, J.~Pool, J.~Tran, and W.~Dally, ``Learning both weights and connections
  for efficient neural network,'' \emph{Advances in neural information
  processing systems}, vol.~28, 2015.

\bibitem{han2015deep}
S.~Han, H.~Mao, and W.~J. Dally, ``Deep compression: Compressing deep neural
  networks with pruning, trained quantization and huffman coding,'' \emph{arXiv
  preprint arXiv:1510.00149}, 2015.

\bibitem{long2014transfer}
M.~Long, J.~Wang, G.~Ding, J.~Sun, and P.~S. Yu, ``Transfer joint matching for
  unsupervised domain adaptation,'' in \emph{Proceedings of the IEEE conference
  on computer vision and pattern recognition}, 2014, pp. 1410--1417.

\bibitem{smith2015rolling}
W.~A. Smith and R.~B. Randall, ``Rolling element bearing diagnostics using the
  case western reserve university data: A benchmark study,'' \emph{Mechanical
  systems and signal processing}, vol.~64, pp. 100--131, 2015.

\bibitem{shao2018highly}
S.~Shao, S.~McAleer, R.~Yan, and P.~Baldi, ``Highly accurate machine fault
  diagnosis using deep transfer learning,'' \emph{IEEE Transactions on
  Industrial Informatics}, vol.~15, no.~4, pp. 2446--2455, 2018.

\bibitem{ruiz2015statistical}
C.~Ruiz-C{\'a}rcel, Y.~Cao, D.~Mba, L.~Lao, and R.~Samuel, ``Statistical
  process monitoring of a multiphase flow facility,'' \emph{Control Engineering
  Practice}, vol.~42, pp. 74--88, 2015.

\end{thebibliography}
%

\bibliographystyle{IEEEtran}

\newpage

\vfill

\end{document}